\pdfoutput=1

\documentclass[11pt]{article}

\usepackage{acl}

\usepackage{times}
\usepackage{latexsym}

\usepackage[T1]{fontenc}
\usepackage{multirow}

\usepackage[utf8]{inputenc}

\usepackage{microtype}
\usepackage{booktabs}
\usepackage{todonotes}
\usepackage{xspace}
\usepackage{cleveref}

\crefformat{section}{\S#2#1#3} 
\crefformat{subsection}{\S#2#1#3}
\crefformat{subsubsection}{\S#2#1#3}

\newcommand{\Nystrom}{Nystr\"om\xspace}

\title{Simple Local Attentions Remain Competitive for Long-Context Tasks}

\author{Wenhan Xiong, Barlas Oğuz, Anchit Gupta, Xilun Chen, Diana Liskovich,  \\ {\bf Omer Levy}, {\bf Wen-tau Yih}, {\bf Yashar Mehdad} \\ \\
 Meta AI 
 \\
 {\small{\tt \{xwhan,barlaso,anchit,xilun,dianaml,omerlevy,scottyih,mehdad\}@fb.com}}
 }

\begin{document}
\maketitle

\begin{abstract}

Many NLP tasks require processing long contexts beyond the length limit of existing pretrained models. To scale these models to longer text sequences, many efficient long-range attention variants have been recently proposed. Despite the abundance of research in this direction, it is difficult to gauge the relative effectiveness of these models in practical use cases, \emph{e.g.}, if we apply these models following the pretrain-and-finetune paradigm. 
In this work, we aim to conduct a thorough analysis of these emerging models with large-scale and controlled experiments. For each attention variant, we pretrain large-size models using the same long-doc corpus and then finetune these models for real-world long-context tasks. Our findings reveal pitfalls of a widely-used long-range benchmark and show that the other efficient attentions fail to outperform the simple local-window attention after standard pretraining. Further analysis of local-attention variants suggests that even the commonly used attention-window overlap is not necessary to achieve good downstream results --- using disjoint local attentions, we are able to build a simpler and more efficient long-doc QA model that matches the performance of Longformer~\citep{longformer} with half of its pretraining compute.\footnote{The code to replicate our experiments can be found at \url{https://github.com/pytorch/fairseq/tree/main/examples/xformers}}


\end{abstract}

\section{Introduction}
The quadratic complexity of Transformer architectures makes it prohibitive to apply large state-of-the-art pretrained models to full-length documents. To efficiently handle longer text while still maintaining the capacity of attention-based models, a long list of efficient attention variants have been proposed and many claim to effectively capture long-range dependencies. Typical paradigms of these architecture innovations involve \textit{learnable sparse attention patterns}~\citep{reformer,sinkhorn,routing}, \textit{fixed local patterns}~\cite{longformer,ETC,Bigbird} and \textit{attention matrix approximation methods}~\cite{linformer,performer,nystrom}. While most of these studies have reported numbers on long sequence inputs, they tend to adopt quite different benchmarks. For instance, Reformer~\citep{reformer} is tested on the 64k-chunk enwik8 dataset for unidirectional language modeling; Performer~\citep{performer} reports masked language modeling (MLM) perplexity on the PG-19 book corpus and protein sequences; Linformer~\citep{linformer} reports MLP perplexity with various input length, while most of the documents in their pretrain corpus are short documents.\footnote{Short documents are concatenated to form long sequences.} The divergence of evaluation protocols makes it hard to compare the relative performance of each attention variant and it is also unknown how they perform well in more practical use cases, which typically involve large-scale pretraining and downstream finetuning. 

Other lines of work such as Longformer~\citep{longformer} and ETC~\citep{ETC} conduct experiments on real-world long-context tasks such as long document QA and summarization. These methods only test fixed local attention patterns, \emph{i.e.}, each token can only attend to a small set of nearby tokens. To reduce the pretraining cost, these models are all initialized from the RoBERTa~\citep{roberta} checkpoint\footnote{By extending the position embeddings and reusing all other parameters.} before further long-doc pretraining. While this paradigm is useful to achieve strong downstream performance, it is not ideal for a fair comparison of all available attention mechanisms, since some of the models use different parametrization that is incompatible with the vanilla transformer attention.


A recently proposed benchmark~\citep{LRA}, named long-range arena (LRA), aims to address the lack of unified evaluation with a bundle of long-sequence tasks. 
However, the text-related tasks in this benchmark are either automatically generated or artificially lengthened by enforcing byte-level inputs, making them rather synthetic. With a fixed byte-level vocabulary and pre-specified model size, all models are trained from scratch with the same epoch limit on each dataset. While the evaluation protocol is consistent across architectures, this setup still deviates from the common paradigm of applying Transformer models, \emph{i.e.},
standard tokenization like BPE or wordpiece,
large-scale pretraining followed and task-specific finetuning~\citep{bert}. Thus, an important question yet to be addressed is whether the results on these artificial datasets are indicative of real-world long-context tasks. 

In this work, our goal is to better understand the effectiveness of various attention mechanisms through a systematic study on practical long-context tasks. Instead of only relying on language modeling or synthetic tasks, we test each model under the standard pretraining-and-finetuning paradigm. For a fair comparison, we implement these attentions under a unified framework and test them using the same Transformer architecture\footnote{We only modify the attention calculation within the multi-head attention blocks} used by RoBERTa-large. We pretrain all models using a large corpus that contains mostly long documents and then finetune them on tasks like long-document question answering, full document retrieval, and text classification. Our experiments show the discrepancies between the commonly used LRA benchmark and downstream results (after pretraining). Additionally, our analysis on the best local attention models allows us to further simplify these models and results in a more efficient long-context encoder. More specifically, the key findings of this paper include:

\begin{itemize}
    \item With proper tuning, we find that all the tested models can achieve similar level of performance on the LRA benchmark while their performance diverges significantly on large-scale pretraining and downstream tasks;
    \item In our experiments, the other attention paradigms barely outperform the class of simple local attentions on downstream tasks when using similar pretraining compute;
    \item As a result of our further analysis of the best performing attention variants, we are able to build a long-doc QA model that is on-par with Longformer while being 2x more efficient. 
\end{itemize}

\section{Preliminaries of Tested Attention Variants}
\label{sec:models}

\begin{figure}[t]
\centering
\includegraphics[width=\linewidth]{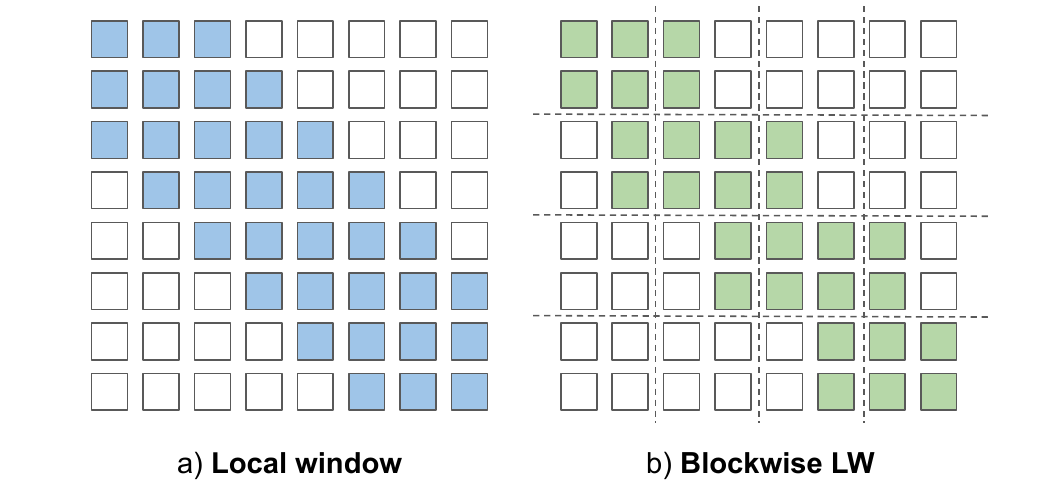}
\caption{Attention pattern visualization of two types of local attentions: \textbf{Left}: Local window attention as in Longformer, with window size 2; \textbf{Right}: Blockwise local window attention with block size 2. The rows represent the tokens in the sequence and the columns represent the tokens being attended to.}
\label{attention_pattern}
\end{figure}

We study three classes of efficient attentions: 

\paragraph{\textit{Fixed local patterns.}} These methods restrict each token to only attend a local window of tokens. The long-range interactions are achieved by the depth of the model. We consider two variants of these models, the token-wise local window attention (\textbf{Local Window}) proposed in \citet{longformer} where each token attends to the same number of tokens on each side, and a simplified and easy-to-implement blockwise version (\textbf{Blockwise LW}) \citep{Bigbird} where each token attends to tokens in the same block and half of the tokens in the left/right blocks. A visualization comparing these two models is shown in Figure \ref{attention_pattern}. 

\paragraph{\textit{Learnable sparse attention patterns.}} Instead of relying on the inductive bias of locality, methods like \textbf{Reformer}~\citep{reformer} and \textbf{Sinkhorn} Attention~\cite{sinkhorn} allow the model to adaptively select tokens to attend to. Briefly, Reformer uses a learnable hashing function to bucket the sequence and each token only attends to tokens in the same bucket; Sinkhorn uses a learnable sorting function to learn a permutation of the segments and each token will attend to tokens in its segment, and the corresponding segment after permutation.

\paragraph{\textit{Kernel-based/Low-rank methods.}} This class of methods use matrix approximation methods to approximate the full attention function. For sequence length $L$ and the hidden dimension $d$, \textbf{Linformer} \citep{linformer} simply uses a projection matrix ($L \times k$) to reduce the length of key and value feature matrix, i.e., from $L \times d$ to $k \times d$ ($k \ll L$). \textbf{\Nystrom} \citep{nystrom} attention adopts a classic matrix approximation method which reconstructs the full attention matrix using a sampled sub-matrix. \textbf{Performer} \citep{performer} eliminates the need of explicitly calculating the $L \times L$ attention matrix by using a random feature method that can approximate the softmax kernel with only dot-product operations. 

\paragraph{\textit{Hybrid attention.}} In addition to these representative methods in each class, our study also includes the more recent \textbf{Long-Short} attention \citep{longshort} which has a similar compression component as in Linformer and combines it with local attentions. Unlike Linformer's compression component which is simply implemented as a standalone projection matrix, \textbf{Long-Short} proposes an input-dependent compression layer, which can adaptively reduce the sequence length.


\paragraph{A note on global tokens.} For many practical NLP tasks, \emph{e.g.},  classification or entailment, the final layer of the model usually requires a single sequence-level representation as input. For local attention models, it is common practice \citep{longformer,Bigbird} to mark a single or a small number of tokens as global tokens and allow these tokens to attend to and be attended by all other tokens. Without incurring much computational cost, these global tokens are important to get better sequence representations and achieve good downstream results. While the mechanism of global tokens has not been used in models with learnable attention patterns, it is straightforward to augment \textbf{Reformer} and \textbf{Sinkhorn} with global tokens using \textit{gather} operations in standard neural network packages, as their attention scores are still calculated by dot product and softmax operations. Thus, in our experiments, except for the kernel-based/low-rank methods, we augment all other models with global tokens to offset the potential performance gap resulting from this trick.

\section{Experiment Setup}

We restrict our studies to encoder-only models and leave the analysis of generative models to future work. We begin by implementing a collection of efficient attentions with a unified framework~\cite{xFormers2021}, which allows us to plug these models into our pretraining-and-finetuning pipeline in a consistent fashion.

\subsection{LRA Experiments} Following recent work on efficient long-range attentions, we take the LRA benchmark as our first set of experiments. As our focus here is on NLP tasks, we consider a subset of LRA tasks with text inputs, \emph{i.e.}, the ListOps, IMDB sentiment analysis, and text matching tasks. All tasks are formulated as classification problems: ListOps requires the model to predict the correct output of an expression (10-way classification), sentiment analysis is to predict the positive/negative labels of IMDB reviews and text matching aims to predict citation links between papers. We follow the hyperparameter settings of recent work~\citep{nystrom,longshort}. Two-layer Transformer encoders are used across all tasks and enough training updates are allowed to ensure convergence\footnote{The limit of training updates is arbitrarily set in LRA and various work have reported hugely improved results on the text matching task, simply by running more training steps.}. Note that this is different from the setup proposed in the original LRA benchmark, where different tasks adopt different model sizes. It is observed from recent work that two-layer models with smaller dimensions are sufficient to achieve similar or better results than previously reported results. The final classification layer is added on top of the representations of \texttt{[CLS]} tokens which are prepended to each sequence. 

\subsection{Pretraining and Downstream Tasks} For practical NLP application, large-scale self-supervised training has become an indispensable ingredient to fully unlock the power of Transformer models. In terms of the experiment scale and testing settings, there is a clear gap between LRA's setup and how we apply state-of-the-art Transformer models in practice. For the second set of experiments, we aim to test these models at scale and investigate whether the results on the LRA benchmark are accurate indicators for real-world long-context tasks after standard large-scale pretraining and finetuning. 

\paragraph{Pretraining Resource.} 
Following \citet{longformer}, we compile a corpus that contains mostly long documents, including Stories~\cite{Stories}, RealNews~\cite{RealNews}, Books corpus~\cite{Books} and English Wikipedia. To make the experiments manageable and relevant for standard GPU hardware, we restrict each model's memory usage close to the 16GB threshold when taking 4,096 tokens in each training batch. We control the batch size and training update across all models:  we use a batch size of 256 sequences ($2^{20}$ tokens) and pretrain each model using the standard masked language modeling objective for 100k updates. We find that all models' training curves almost stabilize after this amount of training steps. We use 32 A100 GPUs for pretraining and all model runs are finished within around 2 days.

\paragraph{Pretraining Architecture} In contrast to Longformer \citep{longformer} and Bigbird \citep{Bigbird} where the models are initialized from RoBERTa before pretraining on long documents, we pretrain these models from scratch, as our goal here is to ensure fair comparison and not all architectures can reuse weights from a standard transformer model. 
In particular, \Nystrom and Performer do not use the standard dot-product and softmax to compute attention probabilities, making their parameters not compatible with common models like RoBERTa or BERT. Furthermore, other models like Linformer or LongShort introduce additional parameters inside the attention module. In our initial experiments, we observe initializing from the RoBERTa put these models at a significant disadvantage compared to other models (e.g., local window attention) that are more compatible with vanilla transformers. Apart from the expanded position embedding matrix and the attention blocks, the architecture hyperparameters are consistent with RoBERTa-large. For both LRA and the large-scale experiments, we adopt the pre layer-normalization trick~\citep{pre_layer_norm} for feedforward and attention blocks. This usually results in better performance in LRA and turns out to be essential for several models in the pretraining experiments.\footnote{Linformer and Performer cannot reach reasonable perplexity without pre-layer normalization.} See additional model-specific architecture settings and models' average memory usage in the Appendix.


\paragraph{Downstream Datasets and Metrics.} We consider practical tasks that naturally involve long documents. We test extractive QA over long documents, long document classification, and document retrieval. For the first two tasks, we use TriviaQA and Hyperpartisan classification respectively, both of which have been used in existing long Transformer work~\cite{longformer}. For full document retrieval, we construct the dataset based on recent open-domain QA work~\citep{odqa} that uses passage-level retrievers. We take an existing passage corpus from \citet{dpr} and reconstruct the document-level corpus. We consider a document to be positive if it includes the answer passage. We reported token-level answer exact match and F1 for extractive QA and the classification accuracy for Hyperpartisan. For the retrieval task, for the ease of experiments, we reported the mean reciprocal rank on the dev set\footnote{For each question, the ground-truth document will be ranked with all documents (both positive and negative) corresponding to the dev-set questions.}, which has been shown to correlate well with final retrieval metric like answer recall \citep{dpr_scale}. We conduct grid search for all tasks and report the best dev results. Given the small size of the Hyperpartisan dataset, we reported averaged results from 4 random seeds. 

\paragraph{Task-specific Architectures for Finetuning.} We use standard architectures for the finetuning tasks: for extractive QA, a single-layer MLP span predictor is added on top of the output token representations; the classification task uses a binary MLP classifier that takes the \texttt{[CLS]} vector as input. For retrieval, we share the query and document encoder using our pretrained models and use dot-product of the \texttt{[CLS]} vectors as the similarity score. For models that are compatible with global tokens, we use all the question tokens as global tokens in the QA task and use a single global token at the start of the sequences for both classification and retrieval. Except for the Hyperpartisan dataset, the document lengths of the other two datasets usually exceed 4,096 tokens after tokenization. In these cases, we drop the tokens outside the models' position range. We put further implementation details and each task's length statistics in the Appendix.

\section{Results and Analysis}
\label{main_results}

\begin{table*}[ht]
\centering
    \begin{tabular}{l|ccc|c|c}
    \toprule
    Model  & ListOps &  Text & Matching & Avg Acc & GFlops\\
    \midrule
    \multicolumn{5}{l}{\emph{Learnable attention pattern}} \\
    Sinkhorn & 37.6 & 63.8 & 80.4 & 60.6 & 0.289\\
    LSH & 37.9 & 62.5 & 80.5 & 60.3 & 0.273\\
    \midrule
    \multicolumn{5}{l}{\emph{Low-rank/kernel-based approximation}} \\
    Linformer & 37.7 & 61.9 & 78.4 & 59.3 & 0.271\\
    Nystrom & 37.9 & 66.1 & 81.0 & 61.7 & 0.256\\
    Performer & 37.1 & 66.1 & 79.8 & 61.0 & 0.205\\
    \midrule
    \multicolumn{5}{l}{\emph{Hybrid attention}} \\
    Long-Short & 37.7 & 65.7 & 81.6 & 61.7 & 0.199\\
    \midrule
    \multicolumn{5}{l}{\emph{Fixed attention pattern}} \\
    Local Window & 37.4 & 65.7 & 81.6 & 61.6 & 0.153 \\
    Blockwise LW & 37.4 & 65.6 & 81.3 & 61.4 & 0.146\\
     \bottomrule
    \end{tabular}
\caption{LRA (the text-input subsets) results with our reimplementations. We did not observe significant performance gaps between different attention variants and simple local attentions remain strong compared to the best \Nystrom attention.}
\label{tab:lra}
\end{table*}

\subsection{Models Perform Similarly in LRA} 
We report our reimplemented LRA results in Table~\ref{tab:lra}. While previous work \citep{LRA} has shown a clear performance gap between different models, we find that with proper tuning, the results of several models could be significantly improved, (\emph{e.g.}, Sinkhorn, Linformer, Reformer, Performer) and \textit{there is no significant performance gap between any of the models} when using a similar level of compute (measure by FLOPS). It is worth noting that these improved results are not obtained by increasing the complexity of models (\emph{e.g.}, by using larger bucket size in Sinkhorn), as our implementation either uses similar or smaller size models compared to existing work.
Also note that while the single global token we added to Sinkhorn and LSH might be essential for some performance gains, it only brings trivial computation overhead. 

\subsection{Pretraining and Downstream Tasks} 
\label{pretrain_and_finetune}



\begin{table*}
\centering
    \begin{tabular}{l|cc|ccc}
    \toprule
    \multirow{2}{*}{Models} & \multicolumn{2}{c|}{MLM Pretraining} & \multicolumn{3}{c}{Downstream Tasks}\\ 
    & PPL $\downarrow$ & k word/sec $\uparrow$ & TriviaQA & Doc Retrieval & Hyperpartisan\\
    \midrule
    \multicolumn{6}{l}{\emph{Learnable attention pattern}} \\
    Sinkhorn     & 4.03 & 11.8 & 63.3/68.5 &  80.9 & 95.0\\
    LSH          & 3.63 & 10.0 & 62.9/67.5 &  83.6 & 92.2\\
    \midrule
    \multicolumn{6}{l}{\emph{Low-rank/kernel-based approximation}} \\
    Linformer       & 4.14        & 24.6   & 59.8/65.2 & 80.3 & 88.7\\
    Nystrom         & 3.79        & 9.5    & 51.5/57.3 & 83.1 & 89.5 \\
    Performer       & 5.58        & 17.2 & 24.5/31.9 & 66.8 & 94.9\\
    \midrule
    \multicolumn{6}{l}{\emph{Hybrid attention}} \\
    Long-Short     & \textbf{3.36} & 8.4 & 66.5/71.4 & 84.5 & 91.5 \\
    \midrule
    \multicolumn{6}{l}{\emph{Fixed local attention pattern}} \\
    Sliding Window  & 3.47        & 9.2 & 65.6/70.7  & 83.2  & \textbf{95.3}\\
    Blockwise LW    & 3.39         & 13.5 & \textbf{68.1/72.9} & \textbf{85.0} & 95.0\\
     \bottomrule
    \end{tabular}
\caption{MLM pretraining and downstream task results.}
\label{tab:pretrain}
\end{table*}

\begin{table}
\small
\centering
    \begin{tabular}{l|c}
    \toprule
    Model & MLM Train Perplexity\\
    \midrule
    Linformer & 4.31 \\
    Performer & 6.36 \\
    Blockwise LW & 4.04 \\
     \bottomrule
    \end{tabular}
\caption{Training perplexity of our best fixed local attention and other faster attention variants. Each model uses similar GPU memory \textbf{and} training time.}
\label{tab:same_pretrain_time}
\end{table}

    

\begin{table*}
\small
\centering
    \begin{tabular}{l|ccccc}
    \toprule
    Model &  MLM PPL & TriviaQA & NQ Doc Retrieval & Hyperpartisan \\
    \midrule
    Blockwise LW & \textbf{3.39} & 68.1/72.9 & 85.0 & 95.0\\
    - w/o overlap & 3.52 & \textbf{68.4/73.2} & \textbf{86.3} &\textbf{96.5}\\
    - w/o overlap \& global tokens & 3.54 & 56.5/61.0 & 85.4 & 94.6\\
    
     \bottomrule
    \end{tabular}
\caption{Ablation of the Blockwise LW Model.}
\label{tab:local}
\end{table*}

We now evaluate these models on practical benchmarks that involve real-world long documents. As shown in Table~\ref{tab:pretrain}, after we scale up the experiments and control the memory consumption of each model, we see more clear differences between these models than what we observe in LRA. Clearly, fixed local attentions remain to be strong baselines. However, in contrast to LRA, we observe local attentions are significantly better than the other attention variants, for both pretraining perplexity and downstream task results. The only exception in terms of the pretraining perplexity is the hybrid Long-Short attention, which already integrates a local attention component: it achieves better perplexity than fixed local attentions, but the downstream results are at most on par with much simpler models like Blockwise LW. It is worth noting that while we only control the training updates and memory usage in Table~\ref{tab:pretrain}, the conclusion still holds if we control the training time of each model: We compare the training perplexity of Blockwise LW attention and other faster models with fixed training time in Table~\ref{tab:same_pretrain_time}.

Even though our LRA experiments also study tasks with text inputs, we see clear discrepancies between the two sets of experiments. Apart from models with fixed local attention patterns, improvements on these text LRA tasks often do not transfer to the standard scaled pretraining-finetuning experiments. For instance, while Performer can outperform most of the non-local attention methods on LRA, it performs poorly on both large-scale MLM and downstream long-context tasks. Similarly, while \Nystrom is significantly better than LSH in LRA on average, we observe the opposite trend in Table~\ref{tab:pretrain}. Among the three tasks, only ListOps is loosely aligned with the MLM perplexity. However, the gaps between each model on this task are still too narrow to be indicative.

Given that large-scale pretraining has become the gold-standard paradigm to build state-of-the-art NLP models. Our findings here call for a more careful and reliable evaluation of lots of existing and emerging long-range attentions. On the other hand, our results also reveal that the local context might still be highly essential even in long context tasks. In the following section, we conduct further analysis on local attention models and attempt to identify the key ingredients of building strong NLP models for downstream long-context tasks.

\subsection{Analysis on Local Attentions}
\label{sec:local_bias}

As we have seen in \cref{pretrain_and_finetune}, models that compute exact attention for local contexts around each token achieve better results. Moreover, the Blockwise LW variant performs the best even it does not guarantee a balanced left and right context window for each token. Given these intriguing findings, we aim to investigate the following questions: \textit{How effective are the long-range mechanism in local attention models?} and \textit{Whether the studied long-context tasks still mostly rely on locality bias?}




\paragraph{Ablation Study.} In the Blockwise LW model, there are two mechanisms that enable long-range connections: the global tokens and the attention window overlap, \emph{i.e.}, each token will additionally attend to half the tokens in the neighboring blocks, and the receptive field increases with model depth. While both are adopted as common practice in existing work~\citep{Bigbird,longformer}, we study the isolated effect of each component in both pretraining and finetuning experiments. For the non-overlap variant, we increase the block size by a factor of 2 such that the amount of tokens each token attends to remains the same. We show the results in Table~\ref{tab:local}. Surprisingly, we see different stories in terms of MLM pretraining and downstream tasks. While both mechanisms are useful for achieving lower MLM perplexity, only the global-token mechanism seems important for downstream tasks. Note that in the document retrieval tasks, removing both mechanisms results in slightly better performance. Now the model is only able to use the first block of the whole document for retrieval. While this seems to suggest that this task is highly local and involves strong positional bias\footnote{The answer context appears at the beginning of the Wikipedia page.}, the gap might be too trivial to be conclusive. Additionally, we only use a single global token for this task, it is likely that assigning more global tokens, \emph{e.g.}, at passage boundaries, could bring additional improvements.
Investigating the particular task further is beyond scope of this work.
In terms of the effect of attention-window overlap, it is expected that this scheme is crucial for lower perplexity: it not only enables more distant dependencies but also reduces the number of "boundary tokens" which can only attend to one side of the context. However, it is counter-intuitive that the overlapping attention links between neighboring blocks, which adds more long-range information, result in worse downstream performance. Also, note that this observation is consistent for all the tasks we studied. There are two possible implications of this finding: 1) the tested tasks still highly depend on locality bias, \emph{i.e.}, most of the important information can be captured solely from the local bias, or 2) the overlapping scheme is not effective at capturing the long-range dependency in downstream tasks. To confirm either hypothesis, we conduct another set of experiments with models that have access to different sizes of context. 

\begin{figure}[t]
\vspace{-0.2in}
\centering
\includegraphics[width=0.9\linewidth]{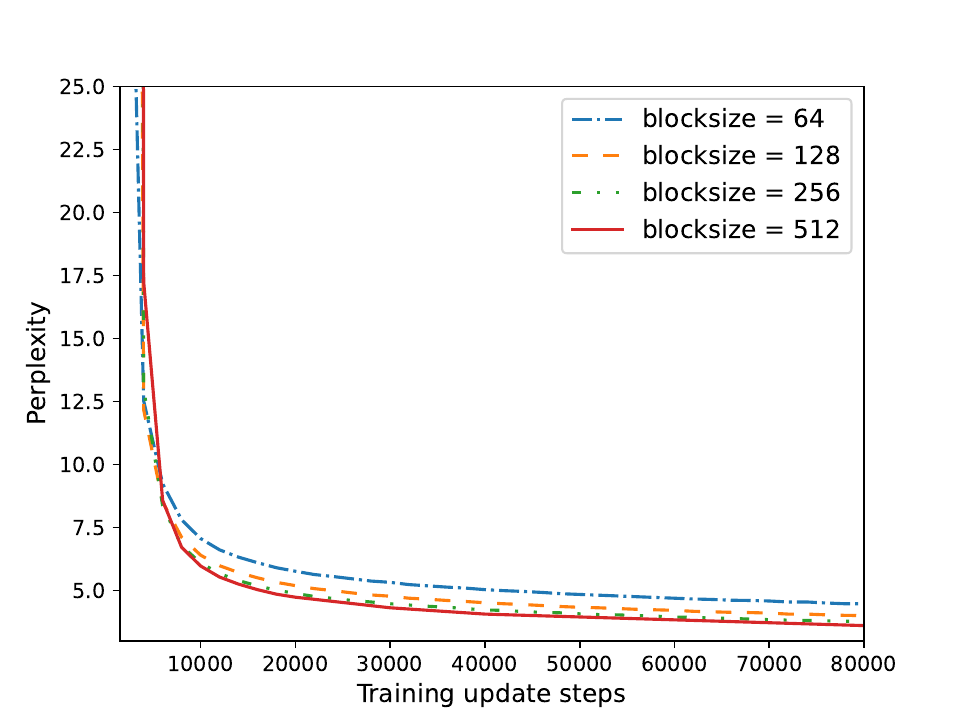}
\caption{Pretraining curves of the non-overlapping block attentions with various context windows.}
\vspace{-0.2in}
\label{curves}
\end{figure}

\paragraph{On Locality Bias.} We take the non-overlapping variant and experiment with various block sizes to see whether longer context is important to studied tasks. We show the results in Table~\ref{tab:block_sizes} and the pretraining curves in Figure~\ref{curves}. While the long-range connections brought by the attention overlap is not helpful for downstream results, we see that increasing the local block sizes does consistently improve both pretraining and downstream performance although the improvement becomes modest beyond block size 256. It is also interesting that the models with smaller block sizes converge faster at the early stage of pretraining. This suggests a staged pretraining process might be more efficient than directly training from long sequences, which aligns with \citet{press-etal-2021-shortformer}'s finding on unidirectional LMs. Overall, this set of experiments suggests that increasing model's capabilities to capture a longer context is generally helpful for both pretraining and downstream tasks. However, using overlapping attention windows is not an effective way to make use of more context. Thus, we hypothesize the MLM perplexity improvements of overlapping local attentions might mainly come from the reduction of the ``boundary" tokens instead of the ability to capture long-range dependencies. For downstream tasks, the issue of ``boundary" tokens is not that essential and the introduction of the overlapping attention windows might disrupt the effective modeling of local context, as the attention module needs to extract both local and distant information from the same set of tokens.\footnote{As the depth of the model increase, the tokens' representation will be added information of more distant tokens.}

\begin{table}[t]
\small
\centering
    \begin{tabular}{l|cc}
    \toprule
    Blocksize & Val PPL & TriviaQA Ans F1\\
    \midrule
    64 & 4.16 & 68.9\\
    128 & 3.74 & 70.7\\
    256 & 3.52 & 73.2\\
    512 & 3.39 & 73.5\\
     \bottomrule
    \end{tabular}
\caption{Pretraining and long-doc QA results of the non-overlapping blockwise attention.}
\label{tab:block_sizes}
\end{table}

\begin{table}
\small
\centering
    \begin{tabular}{l|c|c}
    \toprule
    Blocksize & Speed $\uparrow$   & Ans EM/F1\\
    \midrule
    Longformer (64k) & 6.6k & 73.1/77.8 \\
    \midrule
    Blockwise LW w/o overlap (64k) & 14.8k & \textbf{73.2/77.9}\\
     \bottomrule
    \end{tabular}
\caption{Comparing with Longformer with TriviaQA when initializing the models from RoBERTa. Speed is measure by thousand word per second at pretraining.}
\label{tab:vs_longformer}
\end{table}
\paragraph{Initializing from Existing Short Models.} While we train all models from scratch for the sake of fair comparison,
existing state-of-the-art long context models like Longformer~\citep{longformer} or BigBird~\citep{Bigbird} usually initialize their longer models from an extensively pretrained short model like RoBERTa~\citep{roberta}. With simple techniques like positional embedding copying, a strong long-context encoder can be initialized without the need of pretraining from scratch. To test our findings from the above analysis in this setting, we follow the same scheme but use the non-overlapping block attention as discussed in \cref{sec:local_bias}. We compare this model with Longformer (based on Sliding Window attention) as it uses the same long-doc corpus and pretrain-and-finetune pipeline (\emph{e.g.,} packages and downstream data processing) as our experiments.\footnote{Note that while BigBird has a similar overlapping local attention and outperforms Longformer, it uses a larger corpus, more pretraining compute and different finetune pipelines, making a direct comparison difficult. }
Same as our setting in \cref{pretrain_and_finetune}, here we control the batch size and number of training updates: we use a batch size of 64 and train the model for 64k steps. Note that as we drop the attention window overlaps, the model is 2x more efficient than Longformer: Given the same window/block size $B$ and sequence length $L$, the complexity of the non-overlapping block attention is $L\times B$ compared to Longformer's $2L\times B$. We show the TriviaQA results in Table~\ref{tab:vs_longformer}, where the speed is measured by words per second during pretraining. With only half of the pretraining compute, our model with disjoint attention blocks achieves slightly better performance than Longformer. This confirms that our findings of the attention overlap from the above section are still valid when the models are not trained from scratch.


\section{Related Work}

\paragraph{Long-Range Context in Language Models.}  
Various studies have investigated the effective usage of distant context in unidirectional language models. \citet{lstm_context} look into the context usage of LSTM LMs and find that these models are only capable to make full use of the nearby 50 tokens and the longer range context is only roughly captured, \emph{i.e.}, excluding detailed information such as word orders. Similarly, \citet{transformer_context} studies the mid- and long-range context usage in transformer LMs, by manipulating the ordering and lexical information in the text. Their experiments show that while long-range context is usually helpful, most of the usable information is carried by local ordering statistics and non-function words instead of detailed content like sentence orders. These observations provide a possible explanation of our ablation experiments in \cref{sec:local_bias} that adding overlaps to attention windows does not yield better downstream results, despite allowing the capture of more long-range interaction. \citet{press-etal-2021-shortformer} observe diminishing returns as they increase the context length when using sliding windows at inference time. They propose a staged training paradigm that train LMs from smaller context to longer ones. This paradigm can more efficiently use the training compute and achieves lower perplexity compared to directly training with long sequences. Given that models with smaller attention windows converge faster at early training steps (Figure~\ref{curves}), the staged training might also benefit MLM pretraining but further investigation is required to validate whether it can also bring downstream improvements.

\paragraph{Other Long-Range Architectures.} Instead of modifying the attention calculation, other work proposes to augment transformers with parametric long-term memories. Transformer-XL \citep{transformer_xl} maintains frozen activations of previous tokens in memory and uses them as additional inputs. To handle the shift of positional information of these activations, it also requires a relative position encoding mechanism which brings additional computation cost. The Compressive Transformer \citep{compressive_transformer} takes a similar scheme but proposes to use compression modules to account for even further memories. Both methods cannot be directly applied to long-context understanding tasks. Under the scheme of kernel-based methods, \citet{linear_attention,random_feature_attention,fwp} also attempt to linearize the softmax with kernel methods. The core ideas of these methods are similar to Performer and they only differ in the choice of kernel functions. Outside of the transformer families, a recent work~\citep{sru+} proposes to augment recurrent LMs with minimal attention blocks. It is more efficient while achieving stronger LM perplexity compared to Transformer-XL. However, it is still unknown whether this model scales as well as transformer architectures. 

\section{Conclusion}
We present a systematic study of recent proposed efficient attention variants on real-world long-context NLP tasks. In contrast to existing work, we are the first to test these models with a set of unified and large-scale experiments. Our results reveal the gap between a widely used benchmark and practical downstream tasks after conducting large-scale pretraining. Among all the studied attention methods, we find that the simplest local attentions outperform other complex attention paradigms on downstream tasks. We also show that existing local-attention models can be further simplified by removing the attention-window overlap, resulting in a faster model that achieves similar or better results. Importantly, our work calls for more careful and practical evaluation protocols while developing long-context NLP models.

\bibliography{anthology,custom}

\begin{thebibliography}{30}
\expandafter\ifx\csname natexlab\endcsname\relax\def\natexlab#1{#1}\fi

\bibitem[{Ainslie et~al.(2020)Ainslie, Onta{\~{n}}{\'{o}}n, Alberti, Cvicek,
  Fisher, Pham, Ravula, Sanghai, Wang, and Yang}]{ETC}
Joshua Ainslie, Santiago Onta{\~{n}}{\'{o}}n, Chris Alberti, Vaclav Cvicek,
  Zachary Fisher, Philip Pham, Anirudh Ravula, Sumit Sanghai, Qifan Wang, and
  Li~Yang. 2020.
\newblock {ETC:} encoding long and structured inputs in transformers.
\newblock In \emph{{EMNLP} {(1)}}, pages 268--284. Association for
  Computational Linguistics.

\bibitem[{Beltagy et~al.(2020)Beltagy, Peters, and Cohan}]{longformer}
Iz~Beltagy, Matthew~E. Peters, and Arman Cohan. 2020.
\newblock \href {http://arxiv.org/abs/2004.05150} {Longformer: The
  long-document transformer}.
\newblock abs/2004.05150.

\bibitem[{Choromanski et~al.(2021)Choromanski, Likhosherstov, Dohan, Song,
  Gane, Sarl{\'{o}}s, Hawkins, Davis, Mohiuddin, Kaiser, Belanger, Colwell, and
  Weller}]{performer}
Krzysztof Choromanski, Valerii Likhosherstov, David Dohan, Xingyou Song,
  Andreea Gane, Tam{\'{a}}s Sarl{\'{o}}s, Peter Hawkins, Jared~Quincy Davis,
  Afroz Mohiuddin, Lukasz Kaiser, David~Benjamin Belanger, Lucy~J. Colwell, and
  Adrian Weller. 2021.
\newblock Rethinking attention with performers.
\newblock In \emph{{ICLR}}. OpenReview.net.

\bibitem[{Dai et~al.(2019)Dai, Yang, Yang, Carbonell, Le, and
  Salakhutdinov}]{transformer_xl}
Zihang Dai, Zhilin Yang, Yiming Yang, Jaime Carbonell, Quoc Le, and Ruslan
  Salakhutdinov. 2019.
\newblock \href {https://doi.org/10.18653/v1/P19-1285} {Transformer-{XL}:
  Attentive language models beyond a fixed-length context}.
\newblock In \emph{Proceedings of the 57th Annual Meeting of the Association
  for Computational Linguistics}, pages 2978--2988, Florence, Italy.
  Association for Computational Linguistics.

\bibitem[{Devlin et~al.(2019)Devlin, Chang, Lee, and Toutanova}]{bert}
Jacob Devlin, Ming-Wei Chang, Kenton Lee, and Kristina Toutanova. 2019.
\newblock \href {https://doi.org/10.18653/v1/N19-1423} {{BERT}: Pre-training of
  deep bidirectional transformers for language understanding}.
\newblock In \emph{Proceedings of the 2019 Conference of the North {A}merican
  Chapter of the Association for Computational Linguistics: Human Language
  Technologies, Volume 1 (Long and Short Papers)}, pages 4171--4186,
  Minneapolis, Minnesota. Association for Computational Linguistics.

\bibitem[{Karpukhin et~al.(2020)Karpukhin, Oguz, Min, Lewis, Wu, Edunov, Chen,
  and Yih}]{dpr}
Vladimir Karpukhin, Barlas Oguz, Sewon Min, Patrick Lewis, Ledell Wu, Sergey
  Edunov, Danqi Chen, and Wen-tau Yih. 2020.
\newblock \href {https://doi.org/10.18653/v1/2020.emnlp-main.550} {Dense
  passage retrieval for open-domain question answering}.
\newblock In \emph{Proceedings of the 2020 Conference on Empirical Methods in
  Natural Language Processing (EMNLP)}, pages 6769--6781, Online. Association
  for Computational Linguistics.

\bibitem[{Katharopoulos et~al.(2020)Katharopoulos, Vyas, Pappas, and
  Fleuret}]{linear_attention}
Angelos Katharopoulos, Apoorv Vyas, Nikolaos Pappas, and Fran{\c{c}}ois
  Fleuret. 2020.
\newblock \href {http://proceedings.mlr.press/v119/katharopoulos20a.html}
  {Transformers are rnns: Fast autoregressive transformers with linear
  attention}.
\newblock In \emph{Proceedings of the 37th International Conference on Machine
  Learning, {ICML} 2020, 13-18 July 2020, Virtual Event}, volume 119 of
  \emph{Proceedings of Machine Learning Research}, pages 5156--5165. {PMLR}.

\bibitem[{Khandelwal et~al.(2018)Khandelwal, He, Qi, and
  Jurafsky}]{lstm_context}
Urvashi Khandelwal, He~He, Peng Qi, and Dan Jurafsky. 2018.
\newblock \href {https://doi.org/10.18653/v1/P18-1027} {Sharp nearby, fuzzy far
  away: How neural language models use context}.
\newblock In \emph{Proceedings of the 56th Annual Meeting of the Association
  for Computational Linguistics (Volume 1: Long Papers)}, pages 284--294,
  Melbourne, Australia. Association for Computational Linguistics.

\bibitem[{Kitaev et~al.(2020)Kitaev, Kaiser, and Levskaya}]{reformer}
Nikita Kitaev, Lukasz Kaiser, and Anselm Levskaya. 2020.
\newblock Reformer: The efficient transformer.
\newblock In \emph{{ICLR}}. OpenReview.net.

\bibitem[{Lee et~al.(2019)Lee, Chang, and Toutanova}]{odqa}
Kenton Lee, Ming-Wei Chang, and Kristina Toutanova. 2019.
\newblock \href {https://doi.org/10.18653/v1/P19-1612} {Latent retrieval for
  weakly supervised open domain question answering}.
\newblock In \emph{Proceedings of the 57th Annual Meeting of the Association
  for Computational Linguistics}, pages 6086--6096, Florence, Italy.
  Association for Computational Linguistics.

\bibitem[{Lefaudeux et~al.(2021)Lefaudeux, Massa, Liskovich, Xiong, Caggiano,
  Naren, Xu, Hu, Tintore, and Zhang}]{xFormers2021}
Benjamin Lefaudeux, Francisco Massa, Diana Liskovich, Wenhan Xiong, Vittorio
  Caggiano, Sean Naren, Min Xu, Jieru Hu, Marta Tintore, and Susan Zhang. 2021.
\newblock xformers: A modular and hackable transformer modelling library.
\newblock \url{https://github.com/facebookresearch/xformers}.

\bibitem[{Lei(2021)}]{sru+}
Tao Lei. 2021.
\newblock \href {https://aclanthology.org/2021.emnlp-main.602} {When attention
  meets fast recurrence: Training language models with reduced compute}.
\newblock In \emph{Proceedings of the 2021 Conference on Empirical Methods in
  Natural Language Processing}, pages 7633--7648, Online and Punta Cana,
  Dominican Republic. Association for Computational Linguistics.

\bibitem[{Liu et~al.(2019)Liu, Ott, Goyal, Du, Joshi, Chen, Levy, Lewis,
  Zettlemoyer, and Stoyanov}]{roberta}
Yinhan Liu, Myle Ott, Naman Goyal, Jingfei Du, Mandar Joshi, Danqi Chen, Omer
  Levy, Mike Lewis, Luke Zettlemoyer, and Veselin Stoyanov. 2019.
\newblock \href {http://arxiv.org/abs/1907.11692} {Roberta: {A} robustly
  optimized {BERT} pretraining approach}.
\newblock \emph{CoRR}, abs/1907.11692.

\bibitem[{O{'}Connor and Andreas(2021)}]{transformer_context}
Joe O{'}Connor and Jacob Andreas. 2021.
\newblock \href {https://doi.org/10.18653/v1/2021.acl-long.70} {What context
  features can transformer language models use?}
\newblock In \emph{Proceedings of the 59th Annual Meeting of the Association
  for Computational Linguistics and the 11th International Joint Conference on
  Natural Language Processing (Volume 1: Long Papers)}, pages 851--864, Online.
  Association for Computational Linguistics.

\bibitem[{Oguz et~al.(2021)Oguz, Lakhotia, Gupta, Lewis, Karpukhin, Piktus,
  Chen, Riedel, Yih, Gupta, and Mehdad}]{dpr_scale}
Barlas Oguz, Kushal Lakhotia, Anchit Gupta, Patrick S.~H. Lewis, Vladimir
  Karpukhin, Aleksandra Piktus, Xilun Chen, Sebastian Riedel, Wen{-}tau Yih,
  Sonal Gupta, and Yashar Mehdad. 2021.
\newblock Domain-matched pre-training tasks for dense retrieval.
\newblock \emph{arXiv}, abs/2107.13602.

\bibitem[{Peng et~al.(2021)Peng, Pappas, Yogatama, Schwartz, Smith, and
  Kong}]{random_feature_attention}
Hao Peng, Nikolaos Pappas, Dani Yogatama, Roy Schwartz, Noah~A. Smith, and
  Lingpeng Kong. 2021.
\newblock \href {https://openreview.net/forum?id=QtTKTdVrFBB} {Random feature
  attention}.
\newblock In \emph{9th International Conference on Learning Representations,
  {ICLR} 2021, Virtual Event, Austria, May 3-7, 2021}. OpenReview.net.

\bibitem[{Press et~al.(2021)Press, Smith, and
  Lewis}]{press-etal-2021-shortformer}
Ofir Press, Noah~A. Smith, and Mike Lewis. 2021.
\newblock \href {https://doi.org/10.18653/v1/2021.acl-long.427} {Shortformer:
  Better language modeling using shorter inputs}.
\newblock In \emph{Proceedings of the 59th Annual Meeting of the Association
  for Computational Linguistics and the 11th International Joint Conference on
  Natural Language Processing (Volume 1: Long Papers)}, pages 5493--5505,
  Online. Association for Computational Linguistics.

\bibitem[{Rae et~al.(2020)Rae, Potapenko, Jayakumar, Hillier, and
  Lillicrap}]{compressive_transformer}
Jack~W. Rae, Anna Potapenko, Siddhant~M. Jayakumar, Chloe Hillier, and
  Timothy~P. Lillicrap. 2020.
\newblock \href {https://openreview.net/forum?id=SylKikSYDH} {Compressive
  transformers for long-range sequence modelling}.
\newblock In \emph{8th International Conference on Learning Representations,
  {ICLR} 2020, Addis Ababa, Ethiopia, April 26-30, 2020}. OpenReview.net.

\bibitem[{Roy et~al.(2021)Roy, Saffar, Vaswani, and Grangier}]{routing}
Aurko Roy, Mohammad Saffar, Ashish Vaswani, and David Grangier. 2021.
\newblock \href {https://doi.org/10.1162/tacl_a_00353} {Efficient content-based
  sparse attention with routing transformers}.
\newblock \emph{Transactions of the Association for Computational Linguistics},
  9:53--68.

\bibitem[{Schlag et~al.(2021)Schlag, Irie, and Schmidhuber}]{fwp}
Imanol Schlag, Kazuki Irie, and J{\"{u}}rgen Schmidhuber. 2021.
\newblock \href {http://proceedings.mlr.press/v139/schlag21a.html} {Linear
  transformers are secretly fast weight programmers}.
\newblock In \emph{Proceedings of the 38th International Conference on Machine
  Learning, {ICML} 2021, 18-24 July 2021, Virtual Event}, volume 139 of
  \emph{Proceedings of Machine Learning Research}, pages 9355--9366. {PMLR}.

\bibitem[{Tay et~al.(2020)Tay, Bahri, Yang, Metzler, and Juan}]{sinkhorn}
Yi~Tay, Dara Bahri, Liu Yang, Donald Metzler, and Da{-}Cheng Juan. 2020.
\newblock Sparse sinkhorn attention.
\newblock In \emph{{ICML}}, volume 119 of \emph{Proceedings of Machine Learning
  Research}, pages 9438--9447. {PMLR}.

\bibitem[{Tay et~al.(2021)Tay, Dehghani, Abnar, Shen, Bahri, Pham, Rao, Yang,
  Ruder, and Metzler}]{LRA}
Yi~Tay, Mostafa Dehghani, Samira Abnar, Yikang Shen, Dara Bahri, Philip Pham,
  Jinfeng Rao, Liu Yang, Sebastian Ruder, and Donald Metzler. 2021.
\newblock Long range arena : {A} benchmark for efficient transformers.
\newblock In \emph{{ICLR}}. OpenReview.net.

\bibitem[{Trinh and Le(2018)}]{Stories}
Trieu~H. Trinh and Quoc~V. Le. 2018.
\newblock A simple method for commonsense reasoning.
\newblock abs/1806.02847.

\bibitem[{Wang et~al.(2020)Wang, Li, Khabsa, Fang, and Ma}]{linformer}
Sinong Wang, Belinda~Z. Li, Madian Khabsa, Han Fang, and Hao Ma. 2020.
\newblock Linformer: Self-attention with linear complexity.
\newblock \emph{arXiv}, abs/2006.04768.

\bibitem[{Xiong et~al.(2020)Xiong, Yang, He, Zheng, Zheng, Xing, Zhang, Lan,
  Wang, and Liu}]{pre_layer_norm}
Ruibin Xiong, Yunchang Yang, Di~He, Kai Zheng, Shuxin Zheng, Chen Xing,
  Huishuai Zhang, Yanyan Lan, Liwei Wang, and Tieyan Liu. 2020.
\newblock On layer normalization in the transformer architecture.
\newblock In \emph{International Conference on Machine Learning}, pages
  10524--10533. PMLR.

\bibitem[{Xiong et~al.(2021)Xiong, Zeng, Chakraborty, Tan, Fung, Li, and
  Singh}]{nystrom}
Yunyang Xiong, Zhanpeng Zeng, Rudrasis Chakraborty, Mingxing Tan, Glenn Fung,
  Yin Li, and Vikas Singh. 2021.
\newblock Nystr{\"{o}}mformer: {A} nystr{\"{o}}m-based algorithm for
  approximating self-attention.
\newblock In \emph{{AAAI}}, pages 14138--14148. {AAAI} Press.

\bibitem[{Zaheer et~al.(2020)Zaheer, Guruganesh, Dubey, Ainslie, Alberti,
  Onta{\~{n}}{\'{o}}n, Pham, Ravula, Wang, Yang, and Ahmed}]{Bigbird}
Manzil Zaheer, Guru Guruganesh, Kumar~Avinava Dubey, Joshua Ainslie, Chris
  Alberti, Santiago Onta{\~{n}}{\'{o}}n, Philip Pham, Anirudh Ravula, Qifan
  Wang, Li~Yang, and Amr Ahmed. 2020.
\newblock Big bird: Transformers for longer sequences.
\newblock In \emph{NeurIPS}.

\bibitem[{Zellers et~al.(2019)Zellers, Holtzman, Rashkin, Bisk, Farhadi,
  Roesner, and Choi}]{RealNews}
Rowan Zellers, Ari Holtzman, Hannah Rashkin, Yonatan Bisk, Ali Farhadi,
  Franziska Roesner, and Yejin Choi. 2019.
\newblock Defending against neural fake news.
\newblock In \emph{NeurIPS}, pages 9051--9062.

\bibitem[{Zhu et~al.(2021)Zhu, Ping, Xiao, Shoeybi, Goldstein, Anandkumar, and
  Catanzaro}]{longshort}
Chen Zhu, Wei Ping, Chaowei Xiao, Mohammad Shoeybi, Tom Goldstein, Anima
  Anandkumar, and Bryan Catanzaro. 2021.
\newblock Long-short transformer: Efficient transformers for language and
  vision.
\newblock \emph{NeurIPS}, abs/2107.02192.

\bibitem[{Zhu et~al.(2015)Zhu, Kiros, Zemel, Salakhutdinov, Urtasun, Torralba,
  and Fidler}]{Books}
Yukun Zhu, Ryan Kiros, Richard~S. Zemel, Ruslan Salakhutdinov, Raquel Urtasun,
  Antonio Torralba, and Sanja Fidler. 2015.
\newblock Aligning books and movies: Towards story-like visual explanations by
  watching movies and reading books.
\newblock In \emph{{ICCV}}, pages 19--27. {IEEE} Computer Society.

\end{thebibliography}
\bibliographystyle{acl_natbib}

\clearpage

\appendix

\section{Appendix}
\label{sec:appendix}

\begin{table}[h]
\small
\centering
    \begin{tabular}{l|l}
    \toprule
    Downstream Task & Hyperparameter Grid\\
    \midrule
    TriviaQA &  learning rate: \textit{1e-5, 3e-5, 5e-6};  \\
     &  warmup ratio: \textit{0\%,  10\%} of total steps; \\
    & random seed: \textit{42, 3, 4321};\\
    & batch size: \textit{32}; \\
    & max epochs: \textit{10} \\
    \midrule
    NQ Doc Retrieval &  learning rate: \textit{1e-5, 5e-6, 3e-5};\\
    & random seed: \textit{42, 3}; \\
    & batch size: \textit{8}; \\
    & max epochs: \textit{10}\\
    \midrule 
    Hyperpartisan & learning rate: \textit{1e-5, 3e-5}; \\
    & random seed: \textit{42, 3, 5, 1992}; \\
    & batch size: \textit{8}; \\
    & max epochs: \textit{40}\\
     \bottomrule
    \end{tabular}
\caption{Hyperparamters of downstream finetuning.}
\label{tab:finetune_hyper}
\end{table}

\begin{table}[h]
\small
\centering
    \begin{tabular}{c|c|c}
    \toprule
    \textbf{TriviaQA} & \textbf{Hyperpartisan} & \textbf{NQ doc Retrieval}\\
    Average|P$_{95\%}$ & Average|P$_{95\%}$ & Average|P$_{95\%}$ \\
    \midrule
     \textit{769.8}|\textit{2,067.0} & \textit{3,333.9}|\textit{11,444.3} & \textit{6,732.9}|\textit{17,493.4} \\
     \bottomrule
    \end{tabular}
\caption{Document length statistics in the tested downstream datasets.}
\label{tab:length}
\end{table}
\paragraph{Downstream Task Details.} On TriviaQA, there are usually multiple matched spans in the document, we train the model to maximize the marginalized probability of all matched spans. The prediction head in the classification task is defined as a 2-layer MLP with \textit{tanh} activations. For the retrieval task, we follow existing passage retrieval methods and use in-batch documents as negative retrieval targets. The loss is simply a cross-entropy loss defined over the scores of all documents in the batch. All the models are finetuned using the Adam optimizer with linear decays. We conduct grid search for all the tested models. The hyperparameters for all the three tasks are shown in Table~\ref{tab:finetune_hyper}. In Table~\ref{tab:length}, we show the average and the 95\% percentile of the document lengths in each dataset. As mentioned in the main text, we drop the tokens exceeding 4,096 tokens.

\paragraph{Pretraining Details.} Our pretraining pipeline is implemented with fairseq\footnote{\url{https://fairseq.readthedocs.io/en/latest/}}. We control the memory usage of each model by adjusting model-specifc hyperparameters. The details in shown in Table~\ref{tab:pretrain_hyper}. Due to different model designs, we are not able to exactly control the memory consumption. However, the tested local attentions typical requires less GPU memory than all the other models.

\begin{table*}[t]
\small
\centering
    \begin{tabular}{l|cl}
    \toprule
    Model & Avg Memory Usage (GB) & Architecture Setting\\
    \midrule
    Sinkhorn & 14.2 & block size: $128$ \\
    LSH & 18.2 & num of hash functions: $4$; chunk size: $16$   \\
    Linformer & 17.2 & compression ratio: $8$ \\
    Nystrom & 16.3 & num of landmarks: $256$; convolution kernel size: $35$; \\
    Performer & 14.2 & random feature dimension: $256$; kernel function: $relu$\\
    Long-Short & 16.3 & block size: $128$; num of landmarks: $32$ \\
    Sliding Window & 15.3 & attention window size: $256$ \\
    Blockwise LW & 15.1 & block size: $128$; overlap: $64$ \\
    \midrule 
    Blockwise LW w/o global toks & 14.7 & block size: $128$\\
    Blockwise LW w/o overlap & 13.4 & block size: $256$\\
    Blockwise LW w/o overlap \& global & 13.2 & block size: $256$\\
     \bottomrule
    \end{tabular}
\caption{Model-specific architecture settings and each model's GPU memory usage when feeding in a single sequence of 4,096 tokens.}
\label{tab:pretrain_hyper}
\end{table*}

\end{document}